\documentclass[onecolumn,10pt,cleanfoot]{asme2ej}

\usepackage{graphicx} 
\usepackage{hyperref}   
\hypersetup{
	colorlinks=true,
	linkcolor=blue,
	citecolor=blue,
	urlcolor=blue,
}
\usepackage[square,numbers]{natbib}
\usepackage{algorithm}
\usepackage{algorithmic}
\usepackage{amsmath}
\usepackage{amssymb}
\usepackage{multirow}
\usepackage{overpic}
\usepackage{float}
\newcommand{\argmin}{\operatornamewithlimits{argmin}}
\title{Efficient Propagation of Uncertainty via Reordering Monte Carlo Samples}

\author{Danial Khatamsaz
    \thanks{Corresponding author.} 
    \affiliation{
	Dept. of Mechanical Engineering.\\
	Texas A\&M University\\
	College Station, TX, 77843\\
    Email: khatamsaz@tamu.edu
    }	
}

\author{Vahid Attari
    \affiliation{
	Dept. of Materials Science and Engineering.\\
	Texas A\&M University\\
	College Station, TX, 77843\\
    Email: attari.v@tamu.edu
    }	
}

\author{Raymundo Arr{\'o}yave
    \affiliation{
	Dept. of Materials Science and Engineering.\\
	Texas A\&M University\\
	College Station, TX, 77843\\
    Email: rarroyave@tamu.edu
    }	
}

\author{Douglas L. Allaire
    \affiliation{
	Dept. of Mechanical Engineering.\\
	Texas A\&M University\\
	College Station, TX, 77843\\
    Email: dallaire@tamu.edu
    }	
}

\begin{document}

\maketitle    

\begin{abstract}
{\it 
Uncertainty analysis in the outcomes of model predictions is a key element in decision-based material design to establish confidence in the models and evaluate the fidelity of models.  Uncertainty Propagation (UP) is a technique to determine model output uncertainties based on the uncertainty in its input variables. The most common and simplest approach to propagate the uncertainty from a model inputs to its outputs is by feeding a large number of samples to the model, known as Monte Carlo (MC) simulation which requires exhaustive sampling from the input variable distributions. However, MC simulations are impractical when models are computationally expensive. In this work, we investigate the hypothesis that while all samples are useful on average, some samples must be more useful than others. Thus, reordering MC samples and propagating more useful samples can lead to enhanced convergence in statistics of interest earlier and thus, reducing the computational burden of UP process. Here, we introduce a methodology to adaptively reorder MC samples and show how it results in reduction of computational expense of UP processes.
}
\end{abstract}

\section{Introduction}

In many engineering applications, decision-making processes rely on numerical simulation models. Most often, inputs to numerical models have some sort of uncertainty that induce uncertainty in model outputs. Thus, characterization, propagation, and analysis of uncertainty is a crucial step in any model development task. Understanding uncertainties enables providing a confidence measure to evaluate the applicability of different computational models for decision-making. Uncertainty quantification (UQ) and uncertainty propagation (UP) are recognized as essential components in many engineering applications where UQ refers to understanding uncertainty sources and UP refers to determining output uncertainty of a model due to uncertainties of input variables.

The most common and simplest approach to propagate the uncertainty from input to output is by feeding a large number of inputs to numerical models, known as Monte Carlo (MC) simulation. Based on the strong law of large numbers and the central limit theorem, convergence in the distribution of a quantity of interest is expected. MC integration methods are known as the gold standard approach to carry out UP~\cite{gelman2004bayesian,swiler2006bayesian}.
However, the computational expense associated with MC simulations makes such methods prohibitive and impractical in many engineering applications. To mitigate the computational burden of MC simulations, other methods have been developed such as importance sampling~\cite{melchers1989importance} and adaptive sampling~\cite{bucher1988adaptive}.
Other approaches to carry out UP are local expansion-based methods~\cite{thoft2012structural} that are weak against large variability of inputs, functional expansion-based methods~\cite{xiu2002modeling}, and numerical integration-based methods~\cite{evans1967application}. Change of probability measure from a desired input distribution is another technique to handle UP problems~\cite{amaral2017optimal,sanghvi2019uncertainty,kloek1978bayesian}. There are different ways to transfer a proposal measure to a target measure and one widely used method is the use of Radon-Nikodym (R-N) derivative~\cite{bourgin2006geometric,shepp1966radon}.
A change of measure based on R-N theory is performed by calculating importance weights using the density ratio of target to proposal densities. Note that although the density ratio cannot be calculated via a closed-form expression when underlying probability distributions are unknown, the R-N theory still applies. Accordingly, a sample-based approach has been proposed in Ref.\cite{sanghvi2019uncertainty}. The idea is to generate a large number of hypercubes different in size all over the input space. The density ratio of target and proposal samples is calculated by counting samples inside each hypercube. Next, a system of linear equations, one equation per hypercube, is solved to obtain the weights.  Via sampling from weighted proposal samples, the empirical distribution of target samples is approximated.
Another approach is proposed in Ref.~\cite{10.1007/s11222-016-9644-3} that works with determinable empirical distribution functions. They calculate importance weights by minimizing the L\textsubscript{2}-norm between a weighted proposal empirical distribution and a target distribution function. Although this approach claims to be effective in high-dimensional and large-scale problems, as many samples occupy the boundaries of high-dimensional spaces, numerical ill-conditioning eventually happens. Although implementing a change of measure method enables efficiency gains by skipping the propagation of target samples to computational models, it requires the availability of previously simulated data using the same model on identical input-output spaces. In scenarios where no such set of data or proposal samples exist, there is no choice but directly propagating target samples through computational models.

In this study, we propose an efficient approach to mitigate the computational burden of MC simulation methods for uncertainty propagation purposes. Assume that there exist a large set of samples yet to be propagated through a computational model to obtain the empirical distribution of the model's outputs. While all samples are important on average, the hypothesis here is some samples can be more useful in representing the empirical distribution of all samples. In other words, the Addition or elimination of a particular sample has an impact on the empirical distribution of all the samples, but this impact is not similar for every sample. Herein, the goal is to determine the importance of samples based on their role in defining the empirical distribution of all the samples. Therefore, it is possible to re-order samples based on their importance to be propagated through a model sequentially. Our approach suggests an efficient use of resources by picking the most informative samples when evaluation of all samples is not practical.

The rest of the paper proceeds as follows. In Sec.~\ref{method}, we introduce the proposed framework to reorder samples of a given set based on their importance in representing the empirical distribution of all samples. Next, in Sec.~\ref{demonstration}, the application is demonstrated on an engineering problem that requires running a computationally expensive simulation model. Finally, in Sec.~\ref{conclusions}, we provide concluding remarks and discuss avenues of future works.

\section{Methodology}
\label{method}
In this section, we discuss our proposed method in detail and provide algorithms for easy implementation of the sequentially optimal sampling concept. The method can be applied to any set of samples regardless of the dimensionality and distribution of samples.

In algorithm~\ref{alg1}, different steps of the method are stated. Assuming that we have available a set of samples S yet to be propagated through a model. We are interested to determine the importance of each sample to re-order samples accordingly. Thus, by sequentially propagating them through a model of interest, we assure once the computational resources are exhausted, we obtain the empirical distribution of a quantity of interest with the most similarity to the case where all samples had been propagated.
The algorithm starts by initializing sets P and R to represent sets of sequentially picked samples and the samples yet to be picked respectively. At every iteration, samples from the set R are temporarily augmented to the set P one by one. The goal is to find the sample that minimizes the dissimilarity between empirical distributions of the temporarily updated set of picked samples and set S. Here, we use the Wasserstein metric for this purpose where \textbf{W} = $[w_1, w_2, ..., w_d]$ is the vector that entry $w_i$ indicates Wasserstein distance between two empirical distributions in $i^{t^h}$ dimension of a $d$ dimensional space. At every iteration,  the minimizer of $||\textrm{\textbf{W}}||_1 := \sum\limits_{i=1}^d w_i$ is picked to be added to the set P and to be removed from the set R. In algorithm~\ref{alg1}, function "Wass" takes samples from both sets and calculates Wasserstein distance. We suggest using the Manhattan distance of Wasserstein metric (L\textsubscript{1}-norm) to calculate Wasserstein distance to avoid the dominance of large Wasserstein distance of a single dimension which may cause diminishing reductions in Wasserstein distances in other dimensions.
We call this technique as ``Adaptive Sampling Method''.
\begin{algorithm}[!ht]
        \caption{Adaptive sampling to re-order planned Monte Carlo samples}\label{alg1}
    \begin{algorithmic}
        \STATE \textbf{given:}\\sample set S=$\{\textrm{s}_1,\textrm{s}_2,...,\textrm{s}_n\}$\\ set of sequentially picked samples P=\{\}\\
        \STATE  R $\longleftarrow$ $\textrm{S}-\textrm{P}$
        \WHILE{$\textrm{R}\neq\varnothing$}
                    \STATE \textrm{s}\textsubscript{picked} = $\argmin_{\textrm{s}_i\in \textrm{R}}$ $||$Wass(S,P+s\textsubscript{i})$||_1$ 
                    \STATE P $\longleftarrow$ P+s\textsubscript{picked}
                    \STATE  R $\longleftarrow$ $\textrm{S}-\textrm{P}$
        \ENDWHILE
    \end{algorithmic}
\end{algorithm}
\vspace{-0.3cm}
By implementing algorithm~\ref{alg1}, assuming the goal is to re-order $n$ samples, the algorithm has to complete $\frac{n(n+1)}{2}-1$ iterations which exponentially increases with the number of samples. In such cases, instead of identifying the best sample at each iteration, it is suggested to look for the best batch of samples to update the set P. Algorithm~\ref{alg2} shows different steps in the batch setting. Considering $b$ as the batch size, $k$ different batches of samples are randomly generated by picking $b$ random samples from the set R. Then, instead of augmenting a single sample, a batch of samples is temporarily augmented to the set P to calculate the Wasserstein distance between the sets S and P. The best batch of samples is determined to update the set P and to be removed from the set R accordingly. 
\begin{algorithm}
        \caption{Adaptive sampling to re-order planned Monte Carlo samples in batch setting}\label{alg2}
    \begin{algorithmic}
        \STATE \textbf{given:}\\sample set S=$\{\textrm{s}_1,\textrm{s}_2,...,\textrm{s}_n\}$\\ set of sequentially picked samples P=\{\}\\batch size $b$\\number of batches to generate $k$
        \STATE  R $\longleftarrow$ $\textrm{S}-\textrm{P}$
        \WHILE{$\textrm{R}\neq\varnothing$}
                    \STATE \textrm{b}\textsubscript{picked} = $\argmin^{i=1:k}_{\textrm{b}_i\subseteq \textrm{R}}$ $||$Wass(S,P+b\textsubscript{i})$||_1$ 
                    \STATE P $\longleftarrow$ P+b\textsubscript{picked}
                    \STATE  R $\longleftarrow$ $\textrm{S}-\textrm{P}$
        \ENDWHILE
    \end{algorithmic}
\end{algorithm}
\vspace{-0.3cm}

\section{Demonstration}
\label{demonstration}

    \begin{figure*}[hb]
        \centering
        \includegraphics[width=0.4\textwidth]{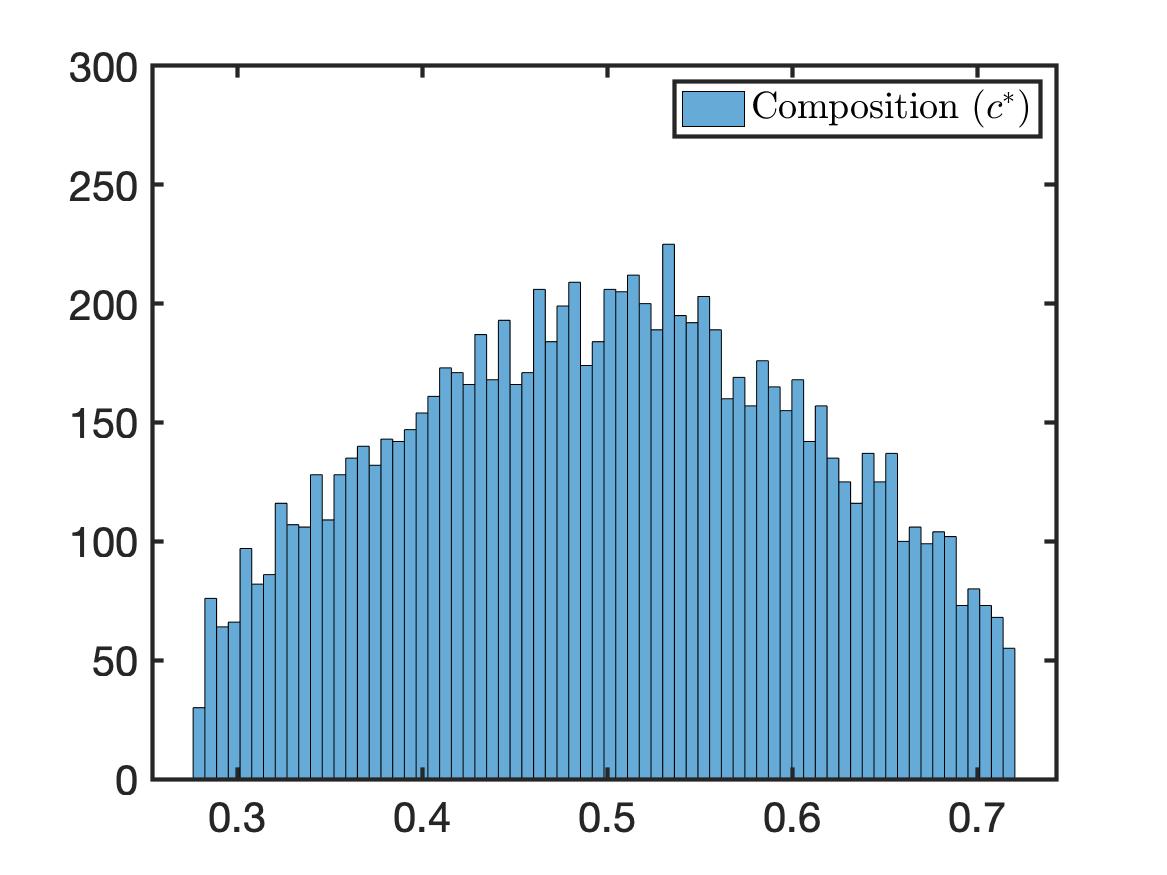}
        \includegraphics[width=0.4\textwidth]{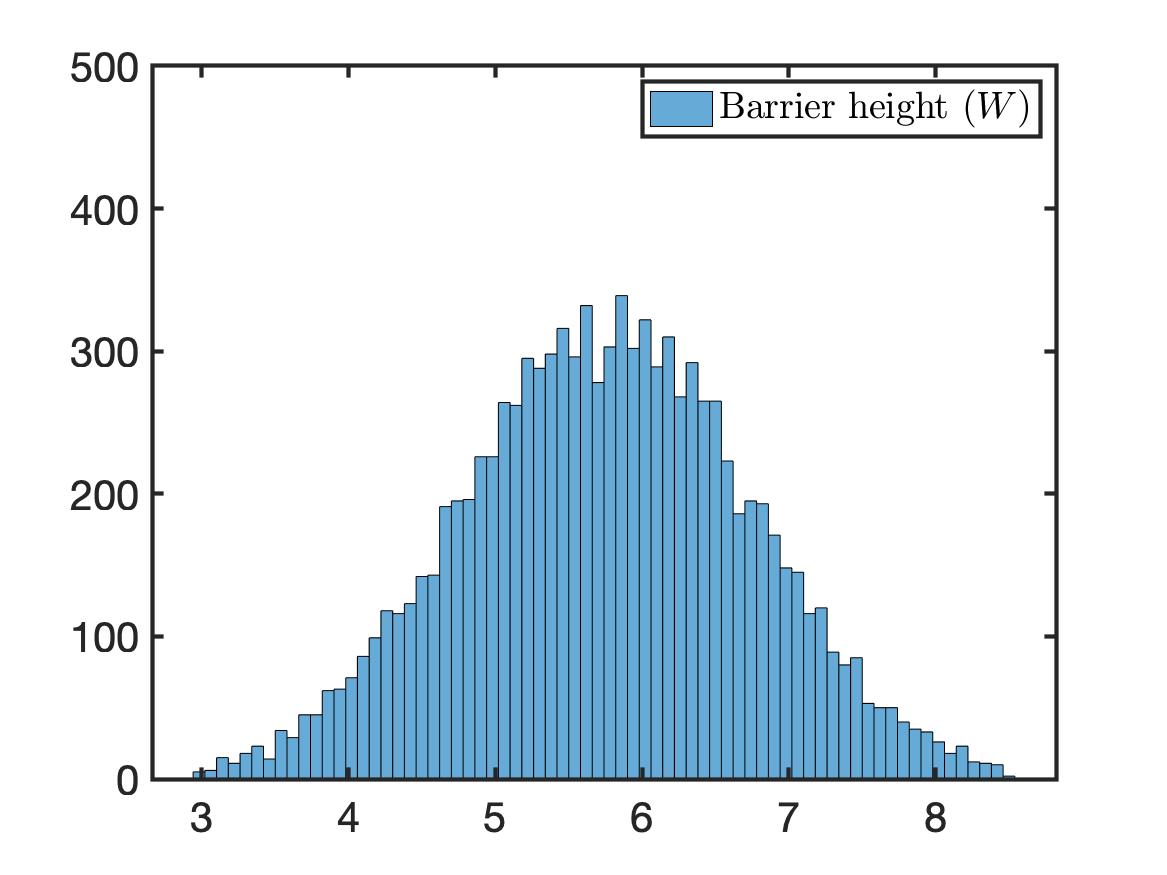}
        \\
        \includegraphics[width=0.4\textwidth]{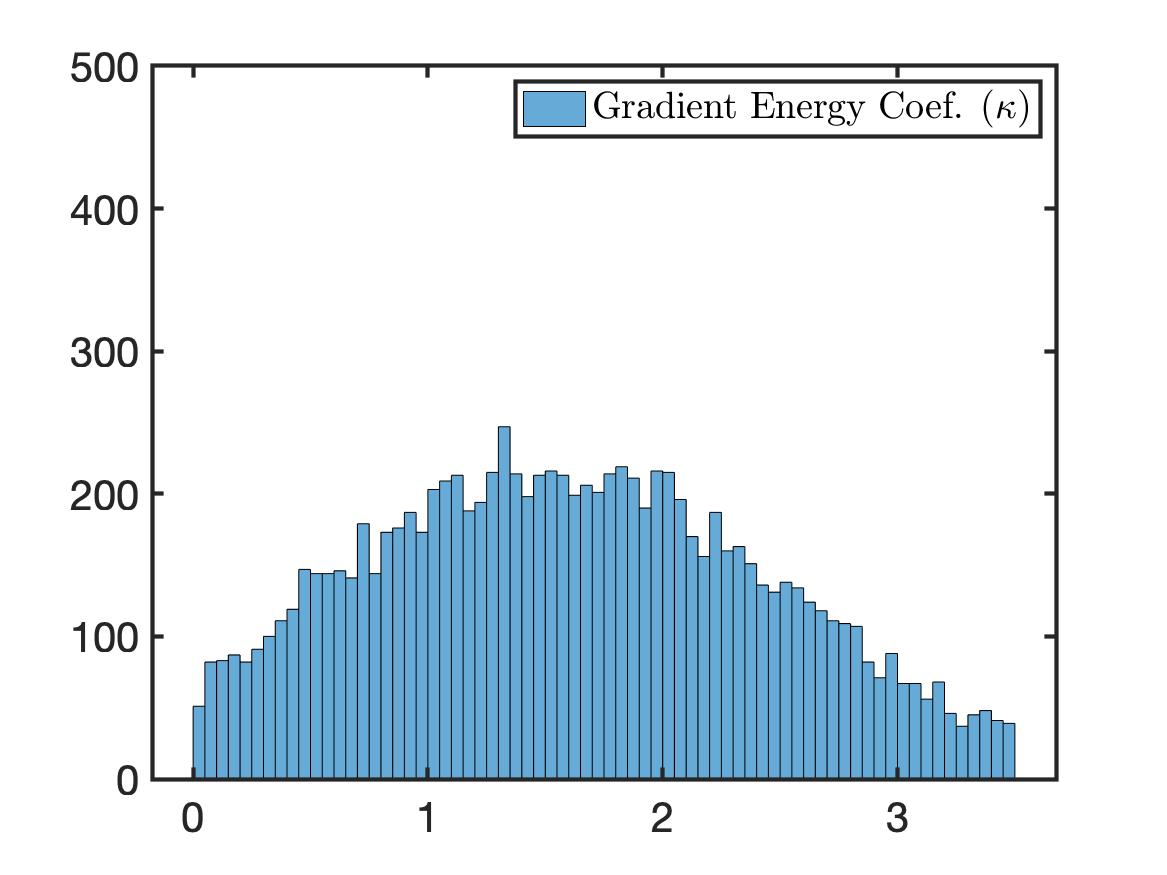}
        \includegraphics[width=0.4\textwidth]{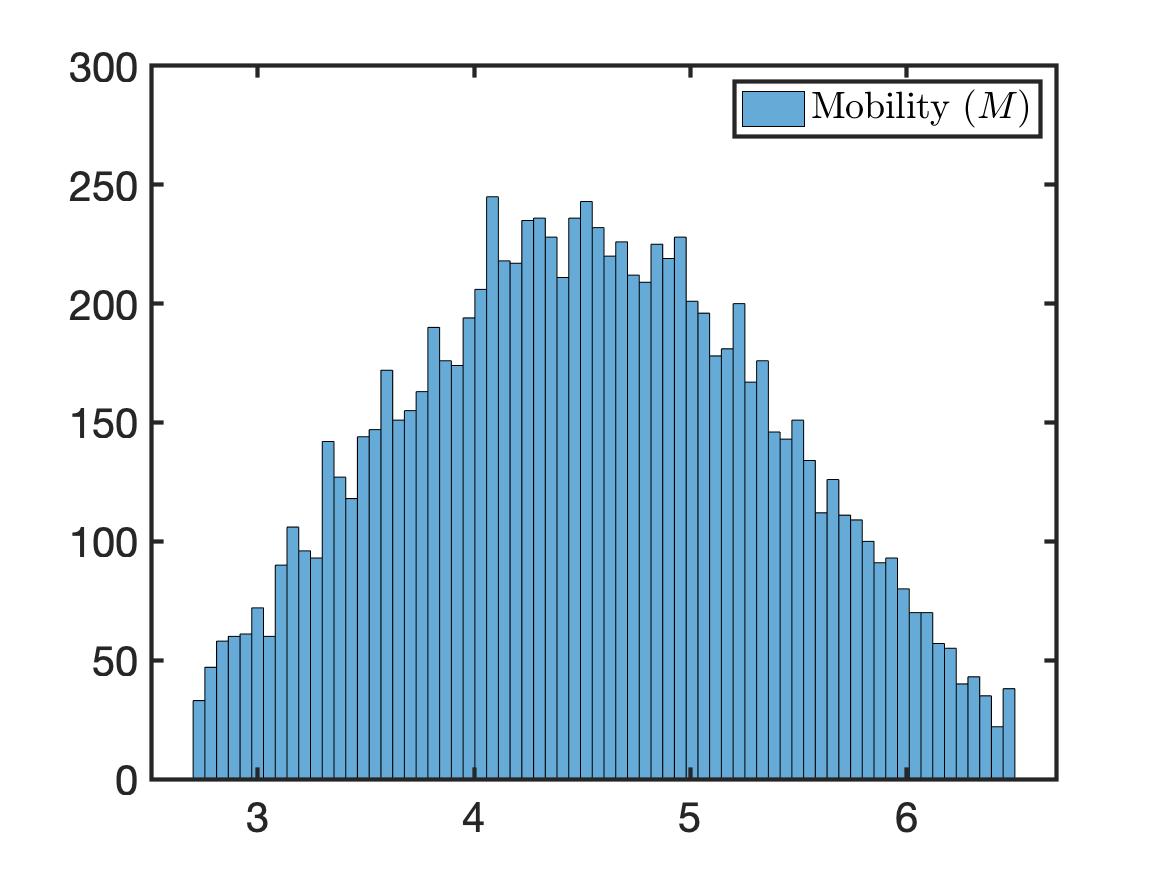}
        \caption{Proposed distributions for input parameters of the phase-field model (i.e., $[c^*,W,\kappa,M]$)}
        \label{fig:parameters_distribution_proposal}
    \end{figure*}
    

    
    A moving boundary problem for the study of interface evolution during spinodal decomposition in alloys is used to demonstrate the framework developed in this study. The model is based on a free energy model for heterogeneous medium accounting for bulk and interfacial free energies,
    \begin{equation}
        F^{tot}(c,\nabla c) = \int_V [f_{bulk} + \frac{\kappa}{2}(\nabla c)^2]dV
    \end{equation}
    where $c$ is alloy composition, $\kappa$ is gradient energy coefficient, and $f_{bulk}$ is the bulk free energy function given as,
    \begin{equation}
         f_{bulk} = W(c-c_{\alpha})(c-c_{\beta})
    \end{equation}
    \noindent
    where $W$ is the barrier height of phase transformation, and $c_{\alpha}$ and $c_{\beta}$ are the equilibrium composition of the product phases that are set to 0.3 and 0.7, respectively. Through high-throughput phase-field simulations, time series of synthetic microstructures will be generated for the investigation of parameter space on the microstructure landscape of a hypothetical alloy during isothermal thermal annealing. The boundary value problem follows:
    \begin{equation}\label{eq:BVP}
        \begin{array}{lccc}
                & \multirow{ 2}{*}{$\dfrac{\partial c}{\partial t}= \nabla . \bigg\{ M \nabla \big( \dfrac{\partial f_{bulk}}{\partial c} - \kappa \nabla^2 c \big) \bigg\}$ } & 0 <x,y< L_x, L_y &  \\ [0pt]
                & &  0<t<t^* & \\ [10pt]
            \text{BC:} & c(0,y,t) = c(L_x,y,t) & c(x,0,t) = c(x,L_y,t) \\
            \text{IC:} & c(x,y,0) = c^* + A\zeta & \\
        \end{array}
    \end{equation}
    \noindent
    where $M$ is the inherently positive effective atomic mobility of the species. The lengths of the simulation domain are set to $L_x=L_y=200$ with the grid size of $256\times256$ and $t^*$ is the final model run time. BC and IC denote the used boundary and initial conditions, respectively. $c^*$ is the initial average value of the alloy composition perturbed by a constant noise magnitude $A$, and $\zeta$ is a Gaussian random number with the interval of $[-1,+1]$. The simulations were carried out using combinations of $[c^*,W,\kappa,M]$ parameter sets. The material properties, such as barrier height of transformation, mobility, and gradient energy coefficient for a given alloy with composition ($c^*$) are often highly uncertain or not available. As a result, a prior distribution with a certain physical range is necessary to be taken into account. Our assumed distributions for these parameters are shown in Fig.~\ref{fig:parameters_distribution_proposal}.

    \begin{figure}[!ht]
        \small
        \centering
        \begin{overpic}[width=0.24\textwidth]{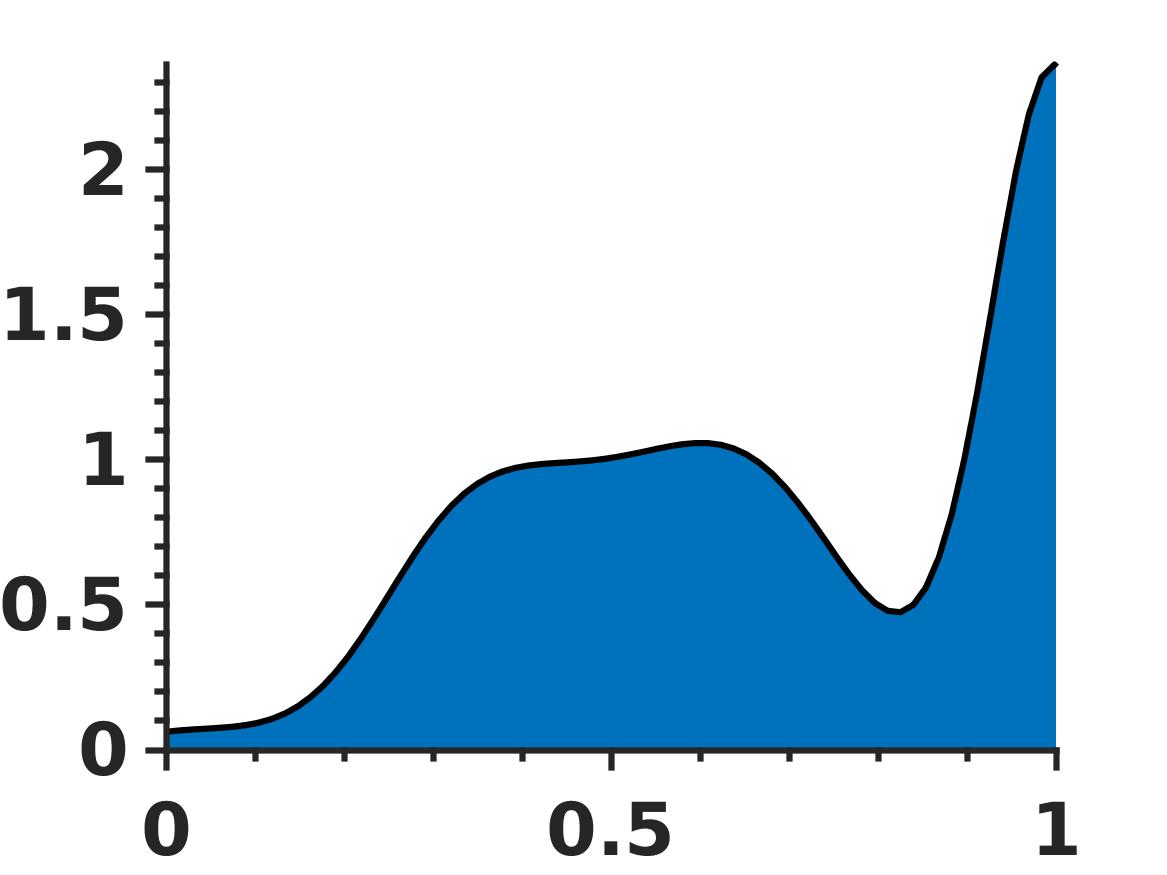}
        \put(20,60){(a) Area fraction}
        \end{overpic}
        \begin{overpic}[width=0.24\textwidth]{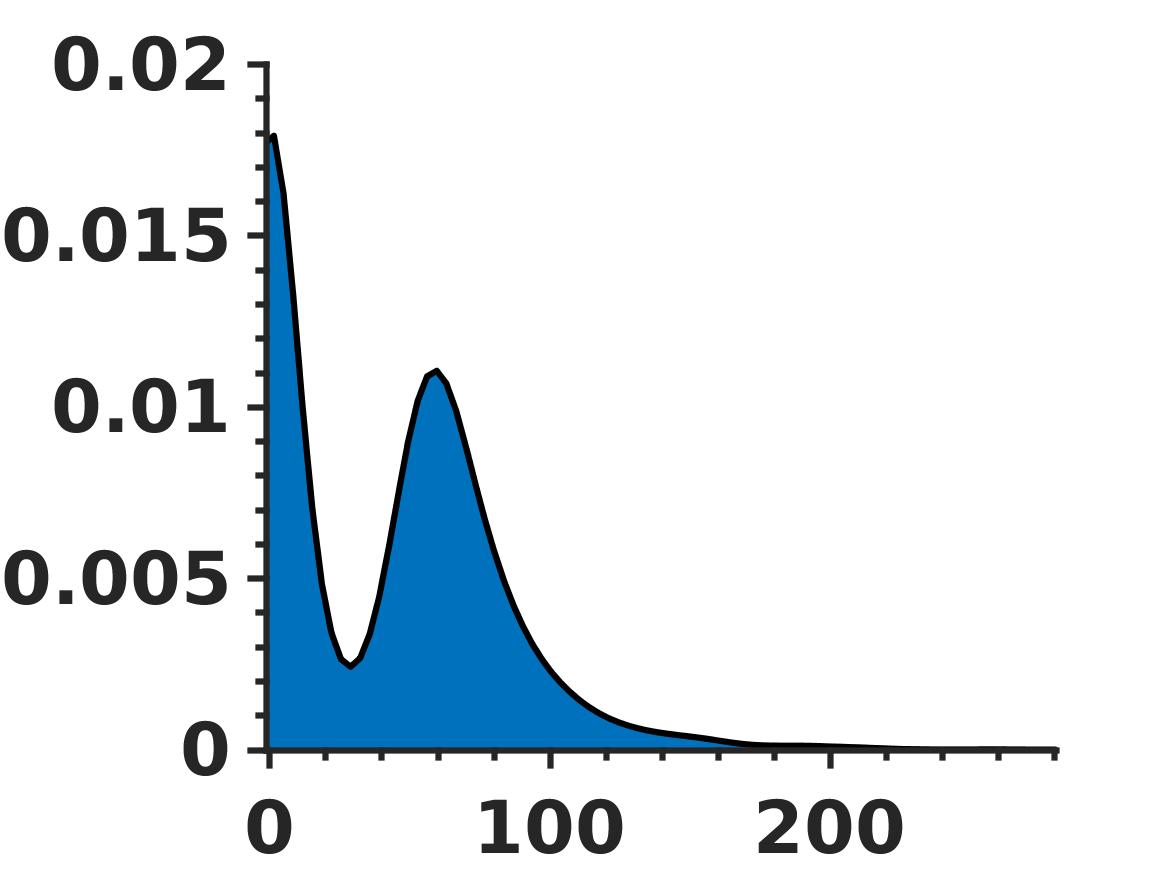}
        \put(27,60){(b) Structure descriptor}
        \end{overpic}
        \normalsize
        \begin{overpic}[width=0.24\textwidth]{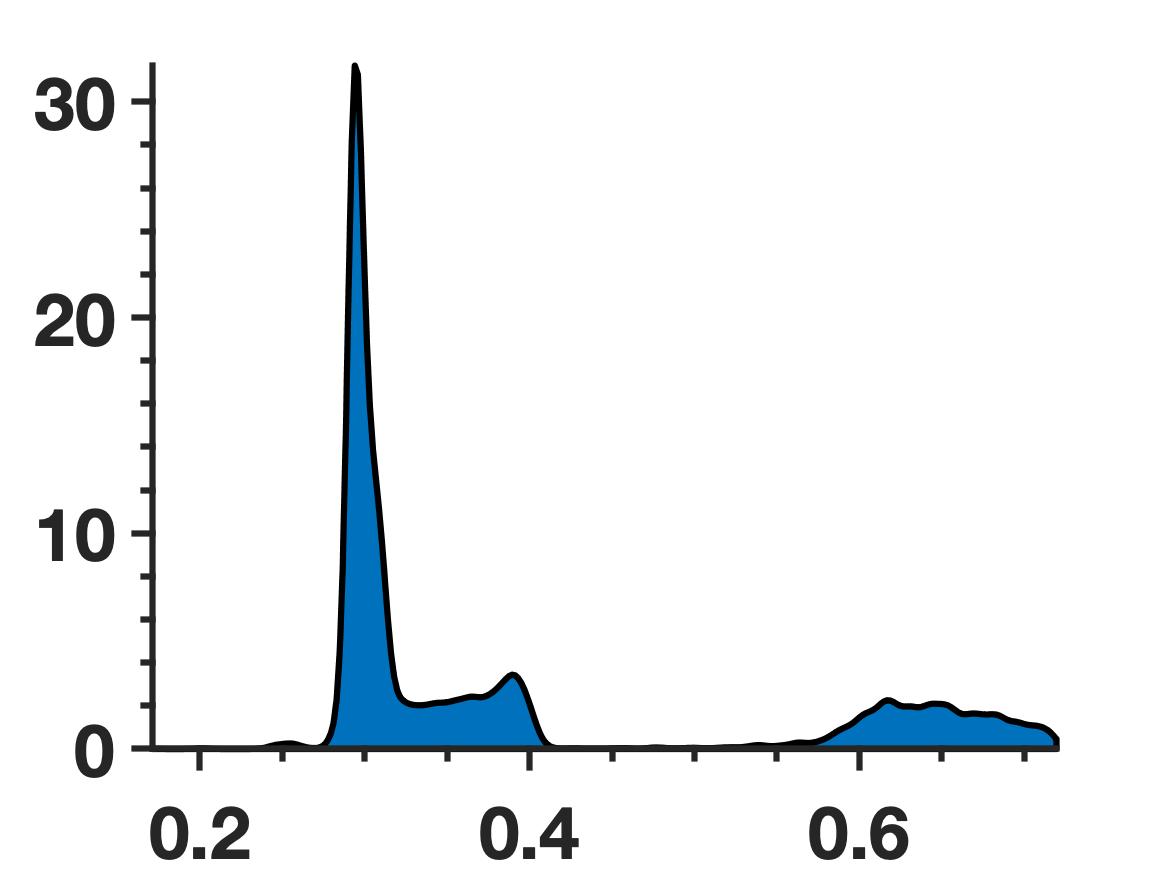}
        \put(40,60){(c) $c_{\alpha}$}
        \end{overpic}        
        \begin{overpic}[width=0.24\textwidth]{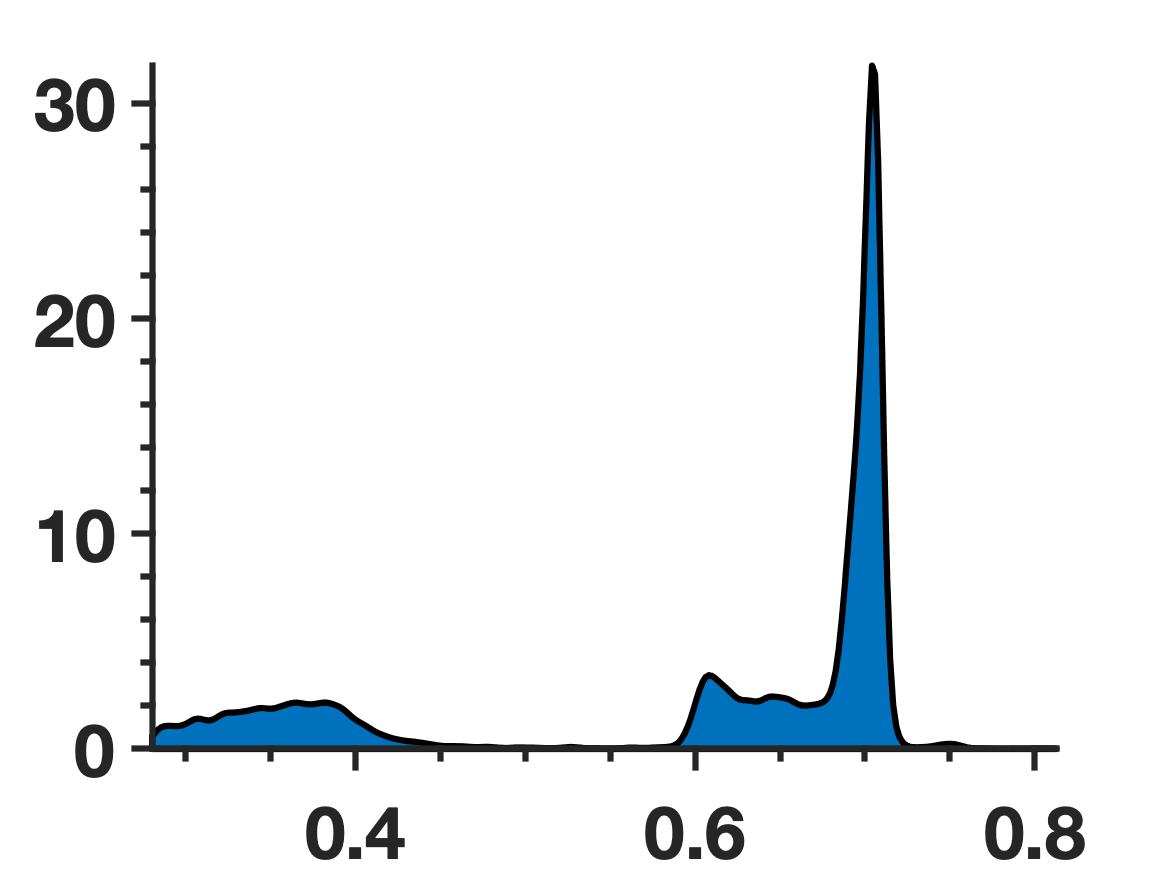}
        \put(20,60){(d) $c_{\beta}$}
        \end{overpic}  
        \vspace{-0.15cm}
        \caption{Probability density functions of Quantities of Interest (QoI) extracted from microstructure images generated by the phase-field model. (a) Area fraction of phases, (b) radially averaged FFT structure descriptor and composition of (c) phase $\alpha$ and (d) phase $\beta$ in the simulation domain}
        \label{fig:QoIs}
    \end{figure}
    
    In this phase-field model, the direct output is time-series images of microstructures, each with a dimension of $256\times256$. For further evaluation of the microstructures, these images are often condensed into a reduced set of physical and non-physical Quantities of Interest (QoI). The conventional reduction of image information is often a one-way transfer without the possibility of inverse transfer from QoI to microstructure image. Due to this condensation, materials' properties and performance are subject to significant uncertainty. Our study determines the area fraction of the phases, the composition of each phase, and the characteristic length scale of the microstructure from radially averaged Fast Fourier Transform (FFT) spectra~\cite{honarmandi2022accelerated}. The probability density functions for these QoIs are extracted for a constant heat treatment duration (i.e., $t^*$) and are shown in Fig.~\ref{fig:QoIs}. These posterior distributions show a diverse range of values for quantities of interest. For instance, the sharp peaks in Fig.~\ref{fig:QoIs}(c) and (d) show the equilibrium composition of 0.3 and 0.7 for the two product phases as dictated by the free energy. Some simulations, however, could also produce non-equilibrium composition values due to an uncertain set of kinetic and thermodynamic parameters. 
\begin{figure}[!hb]
\centering
\includegraphics[width=0.5\columnwidth]{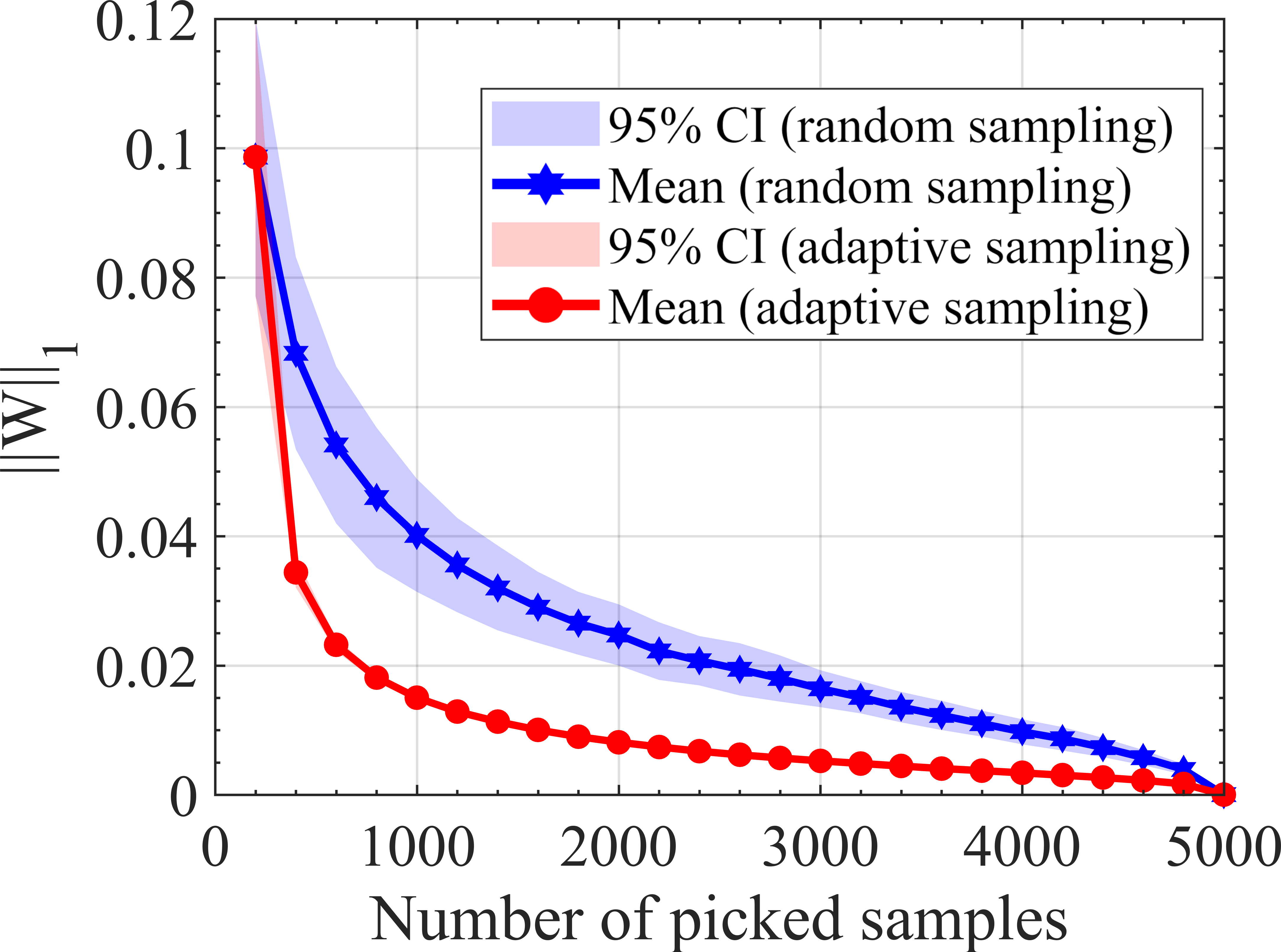}
\caption{Manhattan distance of Wasserstein metric between empirical distributions of sequentially picked samples (adaptive and random sampling) and all samples} 
\label{wass_dist_adaptive}
\end{figure}
Therefore, due to inherent parametric uncertainties, nonlinearity, and difficulties in the post-processing of microstructure images, phase-field simulations are computationally expensive to run. Moreover, advanced phase-field models often combine several order parameters and multiphysics interactions (e.g., thermal, electrical, mechanical, magnetic). This increased degree of complexity in the numerical and parametric calculation of these models often results in the uncertain evaluation of the material's property and performance space during modeling real-world processes (e.g., additive manufacturing~\cite{karayagiz2020finite}, memristive materials for brain-like (neuromorphic) computing~\cite{shi2018phase}, electrodeposition reaction kinetics in battery materials~\cite{davidson2020mapping}, solder interconnect joint formation~\cite{attari2016phase} and electromigration~\cite{attari2018interfacial}, microstructure evolution in thermoelectric materials for energy conversion~\cite{yi2018strain}, to name a few). It is important to note that Eq.~\ref{eq:BVP} (i.e., Cahn-Hilliard equation) and some of its variants are also relevant to phenomena other than phase separation in materials. For instance, tumor growth~\cite{khain2008generalized}, population dynamics~\cite{cohen1981generalized}, image processing~\cite{bertozzi2006inpainting} and even the irregular structure in Saturn's rings~\cite{tremaine2003origin} are some noteworthy examples. 
\begin{figure*}[!ht]
\centering
\includegraphics[width=1\textwidth]{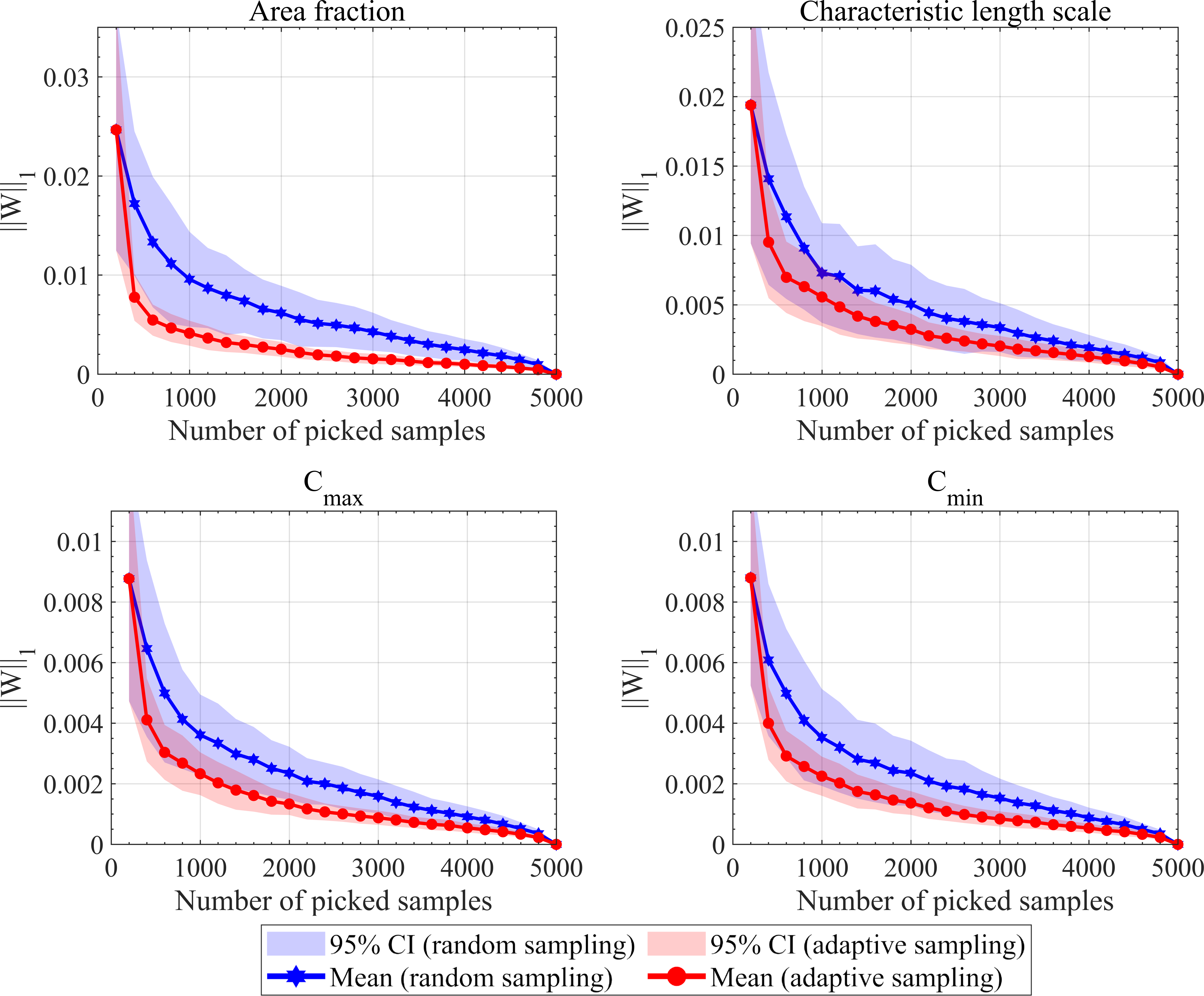}  
\caption{Mean and confidence interval of Wasserstein distance between empirical distributions of sequentially picked samples (adaptive and random sampling) and all samples in the output space}\label{fig:out_qois}
\end{figure*}

We seek to enhance the traditional Monte Carlo sampling methods to enable their use when faced with computationally expensive phase-field models. We, therefore, consider the problem of enhancing the convergence rate of Monte Carlo simulations by creating algorithms that ensure optimal convergence of a sequentially sampled input vectors. 
Here, we have available a set of 5000 samples, and we implement the adaptive sampling method in the batch setting to re-order the samples. We then sequentially propagate them through the model. The Batch size is set to 200 and at every iteration, 20,000 different batches are generated (note that in the last iteration, only 200 samples remain in the set R, so it can be skipped by simply augmenting the last batch to the set P). The simulations are replicated 100 times.
The Manhattan distance of the Wasserstein metric ($\textrm{L}^1$-norm) is plotted in Fig.~\ref{wass_dist_adaptive}. For comparison purposes, the result of a random sampling policy is also illustrated. There are two key points in Fig.~\ref{wass_dist_adaptive}: first, using the adaptive sampling policy to pick the best samples, the Wasserstein distance between the sets of picked samples and all samples is significantly smaller compared to the random sampling policy. This means the same measure of similarity between 2 sets is achieved using a much less number of samples. This emphasizes the fact that some samples are more useful (more informative about the distribution). Thus, based on these results, the same inference about distribution is made using less number of samples if they are picked optimally. The second key point is the narrow confidence interval of the adaptive sampling method. This essentially indicates that almost the same set of samples is consistently recognized at different replications. Note that, multiple replications only apply to the batch setting as at every replication, different sets of batches are generated whereas if the samples are picked one by one, all replication essentially return exactly the same order of samples.

In the next step, samples are propagated through the model to obtain the empirical distributions of all quantities of interest. Figure~\ref{fig:out_qois} illustrates Wasserstein distances comparing the adaptive sampling and random sampling policies. Here, the confidence intervals for both policies are wider since the distances between two samples in the input space and output space are different. However, still significant efficiency gains are observed comparing the required number of propagated samples to achieve the same Wasserstein distance in adaptive and random sampling policies.

\begin{figure}[ht!]
\centering
\includegraphics[width=0.5\columnwidth]{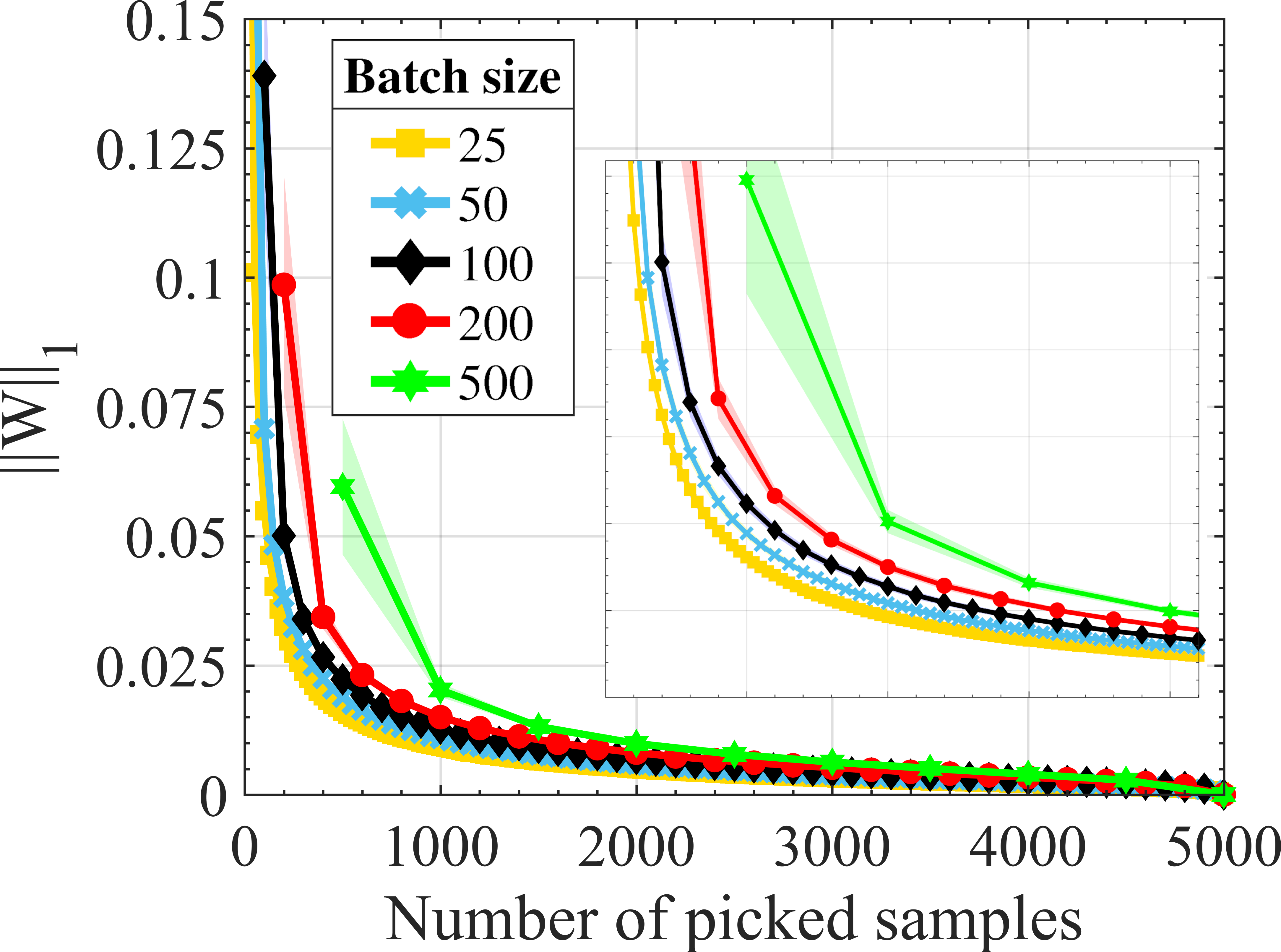}
\caption{Comparison of Wasserstein distances between sequentially picked and all samples at different batch sizes} 
\label{batch_comp}
\end{figure}

As mentioned earlier, one can search for the most useful batch of samples to pick among different generated batches. To investigate the impact of batch size on the performance of the framework, we have performed adaptive sampling using different batch sizes. The results are depicted in Fig.~\ref{batch_comp}.
As we reduce the batch size (increasing resolution), it improves the similarity between the empirical distributions of sequentially picked samples and all samples. However, note that even with the largest batch size, after one iteration, the difference is minimized since the framework recognizes and picks a batch with the most useful samples anyway. Therefore, if the goal is to pick the smallest number of samples possible, a smaller batch size is beneficial while for more relaxed conditions, a larger batch size can also do the job. The best result is achieved when samples are picked sequentially one by one. The trade-off here is smaller batch sizes require more iterations to complete the process. In our problem, since the number of samples is not drastically large (5,000 samples), Wasserstein metric calculations take almost the same computational time at any iteration, thus the wall-time increases almost linearly with respect to the number of iterations.

\section{Conclusions and Future Work}
\label{conclusions}

Although MC simulations suggest a simple approach to propagate uncertainty from a model inputs to its outputs, running thousands of simulations is impractical in many engineering applications. In this work, we introduced the concept of re-ordering MC samples based on their usefulness in representing the empirical distribution of all samples. In this sense, while all samples are important on average, some samples are more informative. We proposed to determine the importance of samples based on their impact on the Wasserstein distance between the sets of all and sequentially picked samples. The more informative a sample is, the more reduction in Wasserstein distance is observed. After re-ordering all samples, they are sequentially propagated through a computational model. The results show significant efficiency gains in comparison to random sample propagation. We also provided the method in the batch setting to decrease computational time by recognizing informative batches of samples instead of testing samples one by one. The results of simulations using different batch sizes suggest that using smaller batch sizes increases efficiency by effectively picking only highly informative samples. However, the batch size effect will be diminished after a few iterations as the framework will pick batches with important samples quickly. 

In this work, the assumption is that samples are generated in the first place and then, we re-order samples before propagating them to computational models. The subject of future work is to propose methods to sample from a distribution consciously instead of generating random samples from input variable distributions. Therefore, efficiency gains are expected in scenarios where sampling from distributions can be computationally demanding.


\begin{acknowledgment}
The authors acknowledge the support of the National Science Foundation through Grant No. CDSE-2001333 and DMR-1905325, as well as ARPA-E through contract DE-AR0001427. Calculations were carried out at the Texas A\&M High-Performance Research Computing (HPRC) Facility.
\end{acknowledgment}

\bibliographystyle{asmems4}
\bibliography{asme2e}

\begin{thebibliography}{10}

\bibitem{gelman2004bayesian}
Gelman, A., Carlin, J.~B., Stern, H.~S., and Rubin, D.~B., 2004,
\newblock ``Bayesian data analysis chapman \& hall,''
\newblock {\em CRC Texts in Statistical Science}.

\bibitem{swiler2006bayesian}
Swiler, L.~P., 2006,
\newblock Bayesian methods in engineering design problems.
\newblock Tech. rep., Sandia National Laboratories (SNL), Albuquerque, NM, and
  Livermore, CA~….

\bibitem{melchers1989importance}
Melchers, R., 1989,
\newblock ``Importance sampling in structural systems,''
\newblock {\em Structural safety, {\bf 6}}(1), pp.~3--10.

\bibitem{bucher1988adaptive}
Bucher, C.~G., 1988,
\newblock ``Adaptive sampling—an iterative fast monte carlo procedure,''
\newblock {\em Structural safety, {\bf 5}}(2), pp.~119--126.

\bibitem{thoft2012structural}
Thoft-Cristensen, P., and Baker, M.~J., 2012,
\newblock {\em Structural reliability theory and its applications}
\newblock Springer Science \& Business Media.

\bibitem{xiu2002modeling}
Xiu, D., and Karniadakis, G.~E., 2002,
\newblock ``Modeling uncertainty in steady state diffusion problems via
  generalized polynomial chaos,''
\newblock {\em Computer methods in applied mechanics and engineering, {\bf
  191}}(43), pp.~4927--4948.

\bibitem{evans1967application}
Evans, D.~H., 1967,
\newblock ``An application of numerical integration techniclues to statistical
  toleraucing,''
\newblock {\em Technometrics, {\bf 9}}(3), pp.~441--456.

\bibitem{amaral2017optimal}
Amaral, S., Allaire, D., and Willcox, K., 2017,
\newblock ``Optimal {L}$_2$ {L}2-norm empirical importance weights for the
  change of probability measure,''
\newblock {\em Statistics and Computing, {\bf 27}}(3), pp.~625--643.

\bibitem{sanghvi2019uncertainty}
Sanghvi, M., Honarmandi, P., Attari, V., Duong, T., Arroyave, R., and Allaire,
  D.~L., 2019,
\newblock ``Uncertainty propagation via probability measure optimized
  importance weights with application to parametric materials models,''
\newblock In AIAA Scitech 2019 forum, p.~0967.

\bibitem{kloek1978bayesian}
Kloek, T., and Van~Dijk, H.~K., 1978,
\newblock ``Bayesian estimates of equation system parameters: an application of
  integration by monte carlo,''
\newblock {\em Econometrica: Journal of the Econometric Society}, pp.~1--19.

\bibitem{bourgin2006geometric}
Bourgin, R.~D., 2006,
\newblock {\em Geometric aspects of convex sets with the Radon-Nikodym
  property}, Vol.~993
\newblock Springer.

\bibitem{shepp1966radon}
Shepp, L.~A., 1966,
\newblock ``Radon-nikodym derivatives of gaussian measures,''
\newblock {\em The Annals of Mathematical Statistics}, pp.~321--354.

\bibitem{10.1007/s11222-016-9644-3}
Amaral, S., Allaire, D., and Willcox, K., 2017,
\newblock ``Optimal l2-norm empirical importance weights for the change of
  probability measure,''
\newblock {\em Statistics and Computing, {\bf 27}}(3), may, p.~625–643.

\bibitem{honarmandi2022accelerated}
Honarmandi, P., Attari, V., and Arroyave, R., 2022,
\newblock ``Accelerated materials design using batch bayesian optimization: A
  case study for solving the inverse problem from materials microstructure to
  process specification,''
\newblock {\em Computational Materials Science, {\bf 210}}, p.~111417.

\bibitem{karayagiz2020finite}
Karayagiz, K., Johnson, L., Seede, R., Attari, V., Zhang, B., Huang, X., Ghosh,
  S., Duong, T., Karaman, I., Elwany, A., et~al., 2020,
\newblock ``Finite interface dissipation phase field modeling of {Ni--Nb} under
  additive manufacturing conditions,''
\newblock {\em Acta Materialia, {\bf 185}}, pp.~320--339.

\bibitem{shi2018phase}
Shi, Y., and Chen, L.-Q., 2018,
\newblock ``Phase-field model of insulator-to-metal transition in {VO$_2$}
  under an electric field,''
\newblock {\em Physical Review Materials, {\bf 2}}(5), p.~053803.

\bibitem{davidson2020mapping}
Davidson, R., Verma, A., Santos, D., Hao, F., Fincher, C.~D., Zhao, D., Attari,
  V., Schofield, P., Van~Buskirk, J., Fraticelli-Cartagena, A., et~al., 2020,
\newblock ``Mapping mechanisms and growth regimes of magnesium
  electrodeposition at high current densities,''
\newblock {\em Materials Horizons, {\bf 7}}(3), pp.~843--854.

\bibitem{attari2016phase}
Attari, V., and Arroyave, R., 2016,
\newblock ``Phase field modeling of joint formation during isothermal
  solidification in {3DIC} micro packaging,''
\newblock {\em Journal of Phase Equilibria and Diffusion, {\bf 37}}(4),
  pp.~469--480.

\bibitem{attari2018interfacial}
Attari, V., Ghosh, S., Duong, T., and Arroyave, R., 2018,
\newblock ``On the interfacial phase growth and vacancy evolution during
  accelerated electromigration in {Cu/Sn/Cu} microjoints,''
\newblock {\em Acta Materialia, {\bf 160}}, pp.~185--198.

\bibitem{yi2018strain}
Yi, S.-i., Attari, V., Jeong, M., Jian, J., Wang, H., Arroyave, R., and Yu, C.,
  2018,
\newblock ``Strain-induced suppression of the miscibility gap in nanostructured
  {Mg2Si}--{Mg2Sn} solid solutions,''
\newblock {\em Journal of Materials Chemistry A}.

\bibitem{khain2008generalized}
Khain, E., and Sander, L.~M., 2008,
\newblock ``Generalized cahn-hilliard equation for biological applications,''
\newblock {\em Physical review E, {\bf 77}}(5), p.~051129.

\bibitem{cohen1981generalized}
Cohen, D.~S., and Murray, J.~D., 1981,
\newblock ``A generalized diffusion model for growth and dispersal in a
  population,''
\newblock {\em Journal of Mathematical Biology, {\bf 12}}(2), pp.~237--249.

\bibitem{bertozzi2006inpainting}
Bertozzi, A.~L., Esedoglu, S., and Gillette, A., 2006,
\newblock ``Inpainting of binary images using the cahn--hilliard equation,''
\newblock {\em IEEE Transactions on image processing, {\bf 16}}(1),
  pp.~285--291.

\bibitem{tremaine2003origin}
Tremaine, S., 2003,
\newblock ``On the origin of irregular structure in saturn's rings,''
\newblock {\em The Astronomical Journal, {\bf 125}}(2), p.~894.

\end{thebibliography}

\end{document}